\documentclass[conference]{IEEEtran}
\IEEEoverridecommandlockouts

\usepackage{cite}
\usepackage{amsmath,amssymb,amsfonts}
\usepackage{algorithmic}
\usepackage{graphicx}
\usepackage{textcomp}
\usepackage[table,xcdraw]{xcolor}
\def\BibTeX{{\rm B\kern-.05em{\sc i\kern-.025em b}\kern-.08em
    T\kern-.1667em\lower.7ex\hbox{E}\kern-.125emX}}
\usepackage{hyperref}

\usepackage{subcaption}

\usepackage{multicol}
\usepackage{tabularx}
\usepackage[flushright, neveradjust]{paralist}

\usepackage{caption}
\usepackage{float}

\usepackage{booktabs}
\usepackage{array}

\usepackage{csquotes}
\newcommand{\q}[1]{\enquote{#1}}

\newenvironment{onsamepage}{
    \noindent\begin{minipage}{\columnwidth}
}{
    \vspace{\parskip}\end{minipage}
}

\begin{document}

\title{Believable Minecraft Settlements\\ by Means of Decentralised Iterative Planning}

\author{
    \IEEEauthorblockN{
        Arthur van der Staaij,
        Jelmer Prins,
        Vincent L. Prins,
        Julian Poelsma,
        Thera Smit, \\
        Matthias M{\"u}ller-Brockhausen,
        Mike Preuss
    }
    \IEEEauthorblockA{
        \textit{LIACS, Universiteit Leiden, The Netherlands} \\
        Correspondence to a.j.w.van.der.staaij@umail.leidenuniv.nl
    }
}

\IEEEoverridecommandlockouts
\IEEEpubid{\begin{minipage}{\textwidth}\ \\[13ex] \centering
© 2023 IEEE. Personal use of this material is permitted. Permission from IEEE must be
obtained for all other uses, in any current or future media, including
reprinting/republishing this material for advertising or promotional purposes, creating new
collective works, for resale or redistribution to servers or lists, or reuse of any copyrighted
component of this work in other works.
\end{minipage}} 
\maketitle
\IEEEpubidadjcol

\begin{abstract}
Procedural city generation that focuses on believability and adaptability to random terrain is a difficult challenge in the field of Procedural Content Generation (PCG). Dozens of researchers compete for a realistic approach in challenges such as the Generative Settlement Design in Minecraft (GDMC), in which our method has won the 2022 competition. This was achieved through a decentralised, iterative planning process that is transferable to similar generation processes that aims to produce ``organic" content procedurally. 
\end{abstract}

\makeatletter
\newcommand*{\rom}[1]{\expandafter\@slowromancap\romannumeral #1@}
\makeatother

\section{Introduction}
\label{sec:intro}

The video game Minecraft has grown to be one of the most-used environments in recent game AI research, probably due to its popularity, flexibility concerning extensions/mods, and its ability to show everything that is happening visually.
There are multiple prominent research directions that use Minecraft as the playing field.
One is to create an agent that controls a player avatar and learns how to make tools in order to eventually obtain a diamond. Project Malmo~\cite{johnson2016the} with the associated NeurIPS competition~\cite{guss2019the} on reinforcement learning in Minecraft have driven this forward, and several algorithms are now known to be able to achieve this goal with decreasing computational effort (sample size). The latest success here is DreamerV3~\cite{hafner2023mastering}.

This work contributes to another avenue of research, one that uses Minecraft as the environment for generative AI.
Procedural Content Generation (PCG) is a vibrant research area in the game AI field that also sees use in the game industry.
Minecraft is an especially well-suited test bed for generative
algorithms, as it is popular, it uses a 3D voxel-based world and it heavily relies on PCG itself for the generation of natural terrain. There are also multiple third-party modifications available that provide the necessary programming interfaces to attach generative algorithms to the game.

\begin{figure}[t]
	\centering
	\includegraphics[width=0.95\columnwidth]{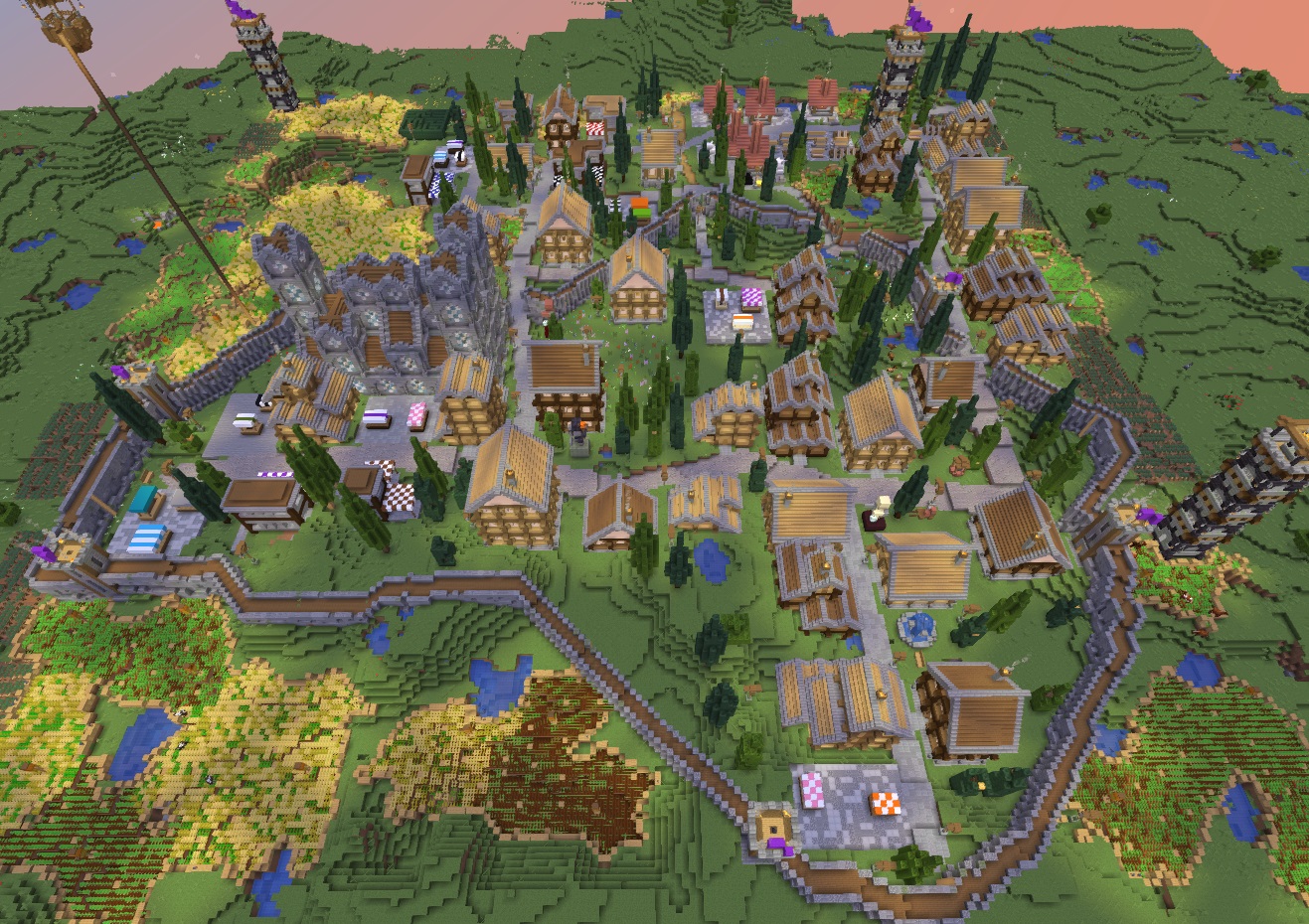}
	\caption{Bird's-eye view of a generated settlement.}
	\label{fig:birdeyesettlement}
\end{figure}

In 2018, Salge et al.\ founded the Generative Design in Minecraft (GDMC) competition\footnote{\url{https://gendesignmc.engineering.nyu.edu/}}~\cite{salge2018generative}, a competition for algorithms that automatically design believable Minecraft settlements that adapt to the natural terrain. Figure~\ref{fig:birdeyesettlement} provides an example of such a generated settlement, created by our algorithm. The competition uniquely employs human judges that judge each entry on four scales: \textit{Adaptatability}, \textit{Functionality}, \textit{Evocative narrative} and \textit{Aesthetics}. The competition has been running annually, which has led to quite a lot of development in this area of research~\cite{green2019organic,fridh2020settlement,9231561,iramanesh2021agentcraft,Esko21,herve2021comparing,9893679,herve2022automated,beukman2023hierarchically}.

Over the years, there has definitely been improvement, with some submissions relying on agent-based solutions
\cite{salge-gdmc-impressions}. It seems that two different understandings of ``agent-based'' exist: one has been used in the sense of NPCs (i.e. avatars) who populate the scene \cite{iramanesh2021agentcraft} and are used to ``model" the inhabitants of a city already during the construction process.
Alternatively, and that is the approach suggested by Lechner~\cite{Lechner2003ProceduralCM, LechnerRWBW06}, agents can be used rather as a means to modularise the building process and enable some decentralised, utility-based decisions.
We follow the latter and extend it in order to better adapt it to the Minecraft competition setup. Our main research question is thus: \emph{How may an iterative agent-based building process as suggested by Lechner be successfully transferred to Minecraft and be extended in order to allow parallel software development?}
Design decisions for this system had to be made in a way such that five human developers could work mostly in parallel, which also means that the resulting system is very flexible concerning future extensions. All that needs to be done is to extend one of the existing agents or make a new one and add it to the collection.

As contributions along the way, we extensively rely on the use of feature maps and a 
bidding system for deciding the land use of a plot. These are general enough to be useful in other generative AI contexts where a complex, \q{organic} looking artefact is designed.

To answer our research question, we first explore the context of existing work on settlement generation in sect.~\ref{sec:related-work}. Thereafter, we present the big picture of our approach in sect.~\ref{sec:overall-approach}. Our realisation of the general approach is discussed in terms of the agents we used in sect.~\ref{sec:agents} and structures in sect.~\ref{sec:structures}. Next, we review the runtime performance of the approach in sect.~\ref{sec:runtime}. After that, we evaluate the quality of our approach based on the scores given by the GDMC jury in sect.~\ref{sec:evaluation}. Instead of a real user study, we report a summary of the jury comments our submission received in sect.~\ref{sec:user-study}. We then develop some ideas for future work in sect.~\ref{sec:future-work} before concluding the paper in sect.~\ref{sec:conclusion}.

\section{Related Work in settlement generation}
\label{sec:related-work}

In \cite{Lechner2003ProceduralCM}, and in more detail in \cite{LechnerRWBW06}, Lechner et al. describe an agent-based system capable of generating cities with three land uses: residential, commercial and industrial land. The system works akin to a game loop: each step/tick, agents get to perform an action. The system also uses agents to construct roads. Lechner's implementation is based on NetLogo, an agent-based simulation framework~\cite{netlogo}.

In their \emph{survey of procedural techniques for city generation}~\cite{kelly2006survey}, Kelly et al. note, however, that this system may \q{be more suitable for simulation applications} rather than procedural generation, due to high computational costs. Nevertheless, one of its benefits is that historical developments can be visualised due to the chronological nature of the agent simulation. Additionally, as the step-based simulation has no definite stopping point, we obtain an anytime algorithm\cite{zilberstein1996using}: stopping the simulation at any step produces a valid result.

However, an agent-based approach may also profitably be employed in a normal imperative programming environment, meaning that we rather import the algorithmic idea but discard the inherently parallel simulation. A first step in this direction has been taken by Esko and Fritiofsson in their thesis~\cite{Esko21}. However, they did not provide many details on how the agents are organised or how they negotiate in case of conflicts.

\section{Overall Approach}
\label{sec:overall-approach}

Our generator has been designed such that several software developers could relatively independently add, modify and remove components without the risk of disabling the building process for the other developers. This is achieved by setting up two important mechanisms for synchronisation:
a common database of \textit{2D feature maps} (2D arrays with the same size of the building area that depict some property) that all further planning processes can rely on, and a bidding mechanism for deciding the use of civilian building plots towards the end.
We may structure the building process into four phases:

\smallskip

\begin{onsamepage}
\begin{compactdesc}
    \item[Terrain analysis phase.] Extracts map data into feature maps
    \item[Blueprint planning phase.] Iteratively increase the size of the planned city 
    \item[Building differentiation Phase.] For each civilian plot, find and apply the best utility agent
    \item[Block placement phase.] Establish the planned structure in Minecraft
\end{compactdesc}
\end{onsamepage}

Steps 2 and 3 can be seen as the \q{simulation} of a city growth process, also referred to as \emph{city planning} or simply \emph{planning}.
Besides these building-related phases, we also enhance the building process with some generation of written narrative. We will describe the building phases and the narrative generation process in more detail below.

\begin{figure}[t]
	\centering
 \vspace*{-1ex}
	\includegraphics[width=1\columnwidth]{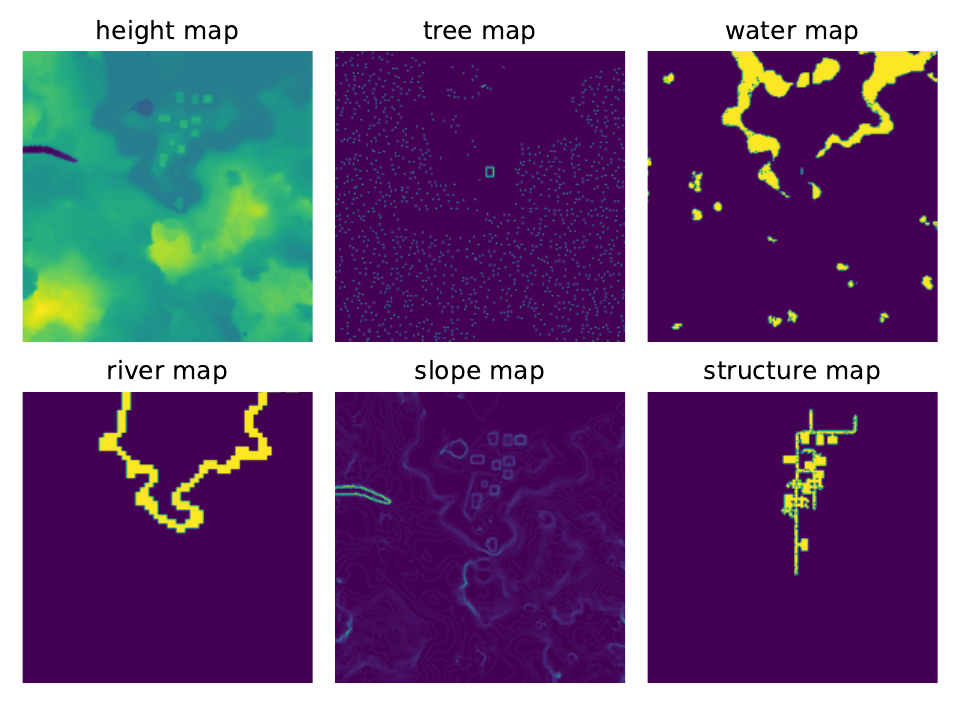}    
	\caption{Terrain-related feature maps generated from a typical 300x300 build area with a pre-existing village.}
	\label{fig:world_data_maps}
\end{figure}

\subsection{Terrain Analysis Phase}
At the beginning of the generation process, we start with a randomly generated Minecraft map that is situated in one or more biomes, has some elevation and water features and some vegetation. Before the blueprint planning phase, the map first has to be analysed in order to inform our agents on possible building locations. We compute several \emph{terrain-related feature maps} that extract useful spatial information relating to the terrain.

Figure~\ref{fig:world_data_maps} shows some of the terrain-related feature maps, of which the most important ones are arguably the height map, the slope map, the water map, and the structure map. They are going to be used and kept track of during the blueprint planning phase in order to decide what can be placed where.
In the heightmap, trees are chopped off and collected in a specific tree~map. The reasoning for this is that, for extending any city, trees can just be removed or planted as fits and are thus not a permanent feature of the landscape.

Besides terrain-related feature maps, we also collect a few qualitative features, such as which wood types appear in the area. These features are used in the block placement phase later on in order to blend in with the environment.

\subsection{Blueprint Planning Phase}
Now that we have the feature maps that give us info on the terrain, we can start simulating the city growth. This is first done purely in-memory in the form of a blueprint, without placing any blocks in the game. This provides some abstraction, since parts can still be removed and we will need the complete blueprint in order to determine how any individual building will look. The planning phase uses a list of agents detailed in sect.~\ref{sec:agents}. Each of these will try to add something to the blueprint, such as roads, houses or trees, based on their preferences and constraints. During the planning phase, we also keep track of another set of \emph{blueprint-related feature maps}, the most important of which are listed in Table~\ref{table:map_types}.
These feature maps change during the planning phase to reflect updates.

\begin{figure}[t]
    \centering
    \small
    \begin{tabularx}{\columnwidth}{c X}
        \toprule
        \textbf{Map type} & \textbf{Explanation \& Example use} \\
        \midrule
        Road& 
            Whether the block contains a road. Useful for placing plots next to roads, but also to avoid colliding with them.\\
        \addlinespace
        Plot& 
            Whether the block is part of an already planned plot. Useful for collision detection. \\
        \addlinespace
        City Wall& 
            Whether the block is part of the city walls. Useful for collision detection. \\
        \addlinespace
        Walled-in& 
            Whether the block is inside the area enclosed by city walls. Useful for buildings built either only outside of city walls (e.g. farms) or only inside city walls (e.g. fountains). \\
        \bottomrule
    \end{tabularx}
    \captionof{table}{A summary of the most important feature maps that the blueprint system keeps track of. All maps are boolean maps with dimensions matching the build area.}\label{table:map_types}
\end{figure}

\begin{figure}[t]
	\centering
 
	\begin{subfigure}[b]{0.32\linewidth}
		\includegraphics[width=\linewidth]{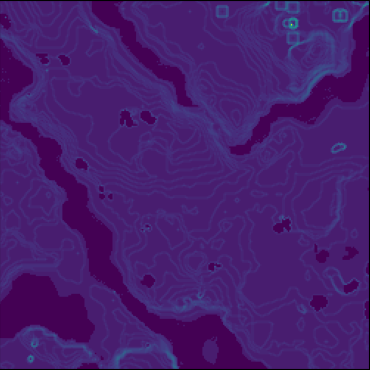}
		\caption{T=0}
	\end{subfigure}
	\hfill
	\begin{subfigure}[b]{0.32\linewidth}
		\includegraphics[width=\linewidth]{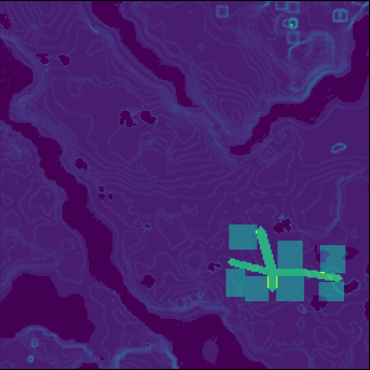}
		\caption{T=5}
	\end{subfigure}
	\hfill
	\begin{subfigure}[b]{0.32\linewidth}
		\includegraphics[width=\linewidth]{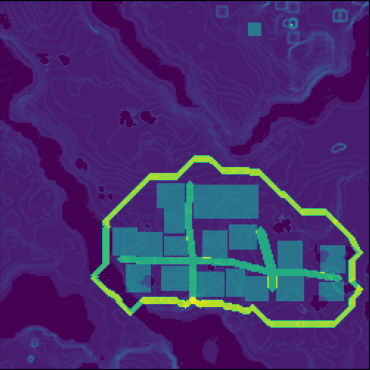}
		\caption{T=10}
	\end{subfigure}
 	\hfill

	\begin{subfigure}[b]{0.32\linewidth}
		\includegraphics[width=\linewidth]{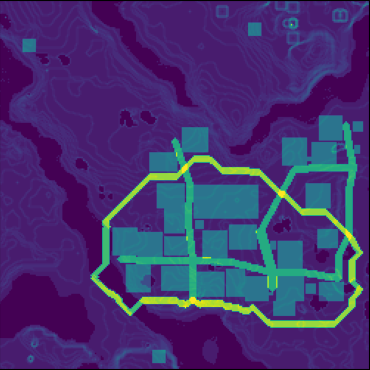}
		\caption{T=15}
	\end{subfigure}
	\hfill
	\begin{subfigure}[b]{0.32\linewidth}
		\includegraphics[width=\linewidth]{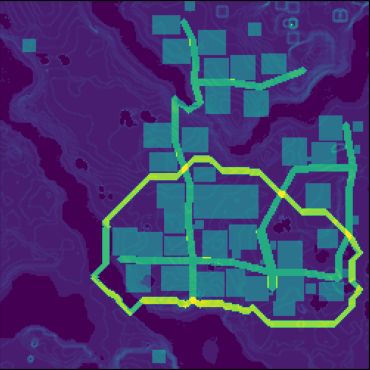}
		\caption{T=20}
	\end{subfigure}
	\hfill
	\begin{subfigure}[b]{0.32\linewidth}
		\includegraphics[width=\linewidth]{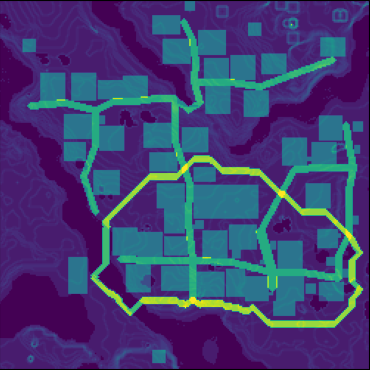}
		\caption{T=25}
	\end{subfigure}
 	\hfill

	\begin{subfigure}[b]{0.32\linewidth}
		\includegraphics[width=\linewidth]{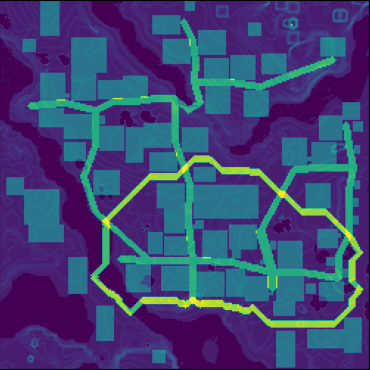}
		\caption{T=30}
	\end{subfigure}
	\hfill
  	\begin{subfigure}[b]{0.32\linewidth}
		\includegraphics[width=\linewidth]{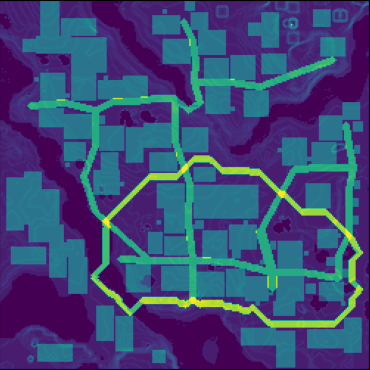}
		\caption{T=45}
	\end{subfigure}
	\hfill
	\begin{subfigure}[b]{0.32\linewidth}
		\includegraphics[width=\linewidth]{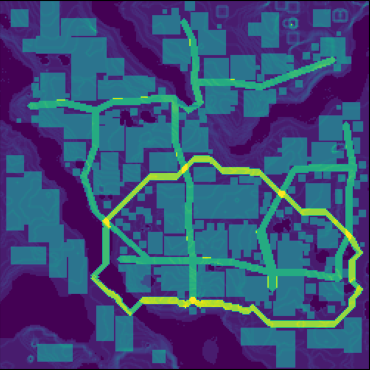}
		\caption{T=60}
	\end{subfigure}
 	\hfill
 
	\caption{Visualisations of the city blueprint at different time steps in the planning phase. When to stop the process is a matter of taste. We use 60 time steps as a typical value.}
	\label{fig:blueprint_evolution}
\end{figure}

At the start of the planning phase, we choose the spot with the flattest terrain and place a small piece of road to kick-start the iterative expansion. After the initial road piece is placed, the step-wise generation process starts, where multiple agents are called to try to add their respective structures to the blueprint. Each agent can either succeed, or fail and skip their turn. The city is thus built up step-by-step in a realistic expansion process. Figure~\ref{fig:blueprint_evolution} visualises this iterative nature of the blueprint. For example, a road agent has to start expanding the road before more houses can be placed, but certain map areas will be too steep for roads, or there may be too much water to cross in the way. Some agents even have the ability to remove blueprint components that have been created by other agents in order to make room for their own structure.
This is, for example, done by the church-placer agent, since we as designers view the church as an important part of a city and want to ensure room is created for its placement.

Agents can be added and removed from the list of active agents at any time. An example use-case for this are tree-planting agents. We want to fill in the gaps between structures with trees towards the end of the blueprint planning phase, but do not want trees to be \q{in the way} during the bulk of the planning phase. We can also manually decide how often agents should run and how often their respective structures are allowed to exist in the world. This gives us some more guiding control over the number of structures and the order in which they are generated.

Each agent has certain constraints on where their structures are allowed to be placed (e.g. boats on water, houses near roads). An agent first finds all locations that match these constraints. Next, a random subset of these locations is evaluated for their fitness and the best is chosen. This fitness function can change depending on the agent. For example, watchtowers use a fitness function that rewards a location that is both on high terrain and far away from other watchtowers.

\begin{figure*}[ht]
	\centering
	\begin{subfigure}[b]{0.48\linewidth}
		\includegraphics[width=\linewidth]{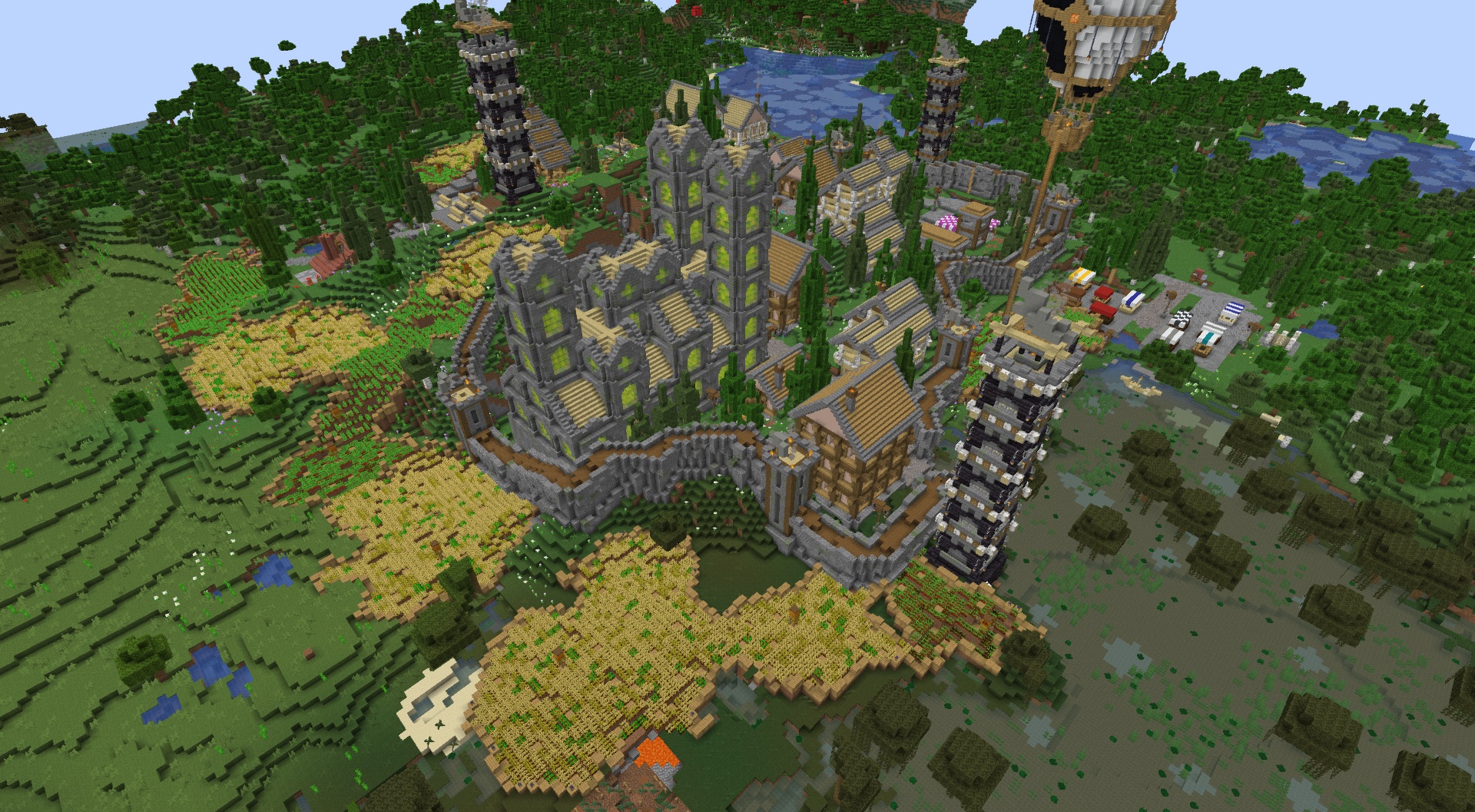}
		\caption{Bird's-eye view of a city in a plains biome.}
        \label{fig:birdseye_plains}
	\end{subfigure}
	\hfill
	\begin{subfigure}[b]{0.48\linewidth}
		\includegraphics[width=\linewidth]{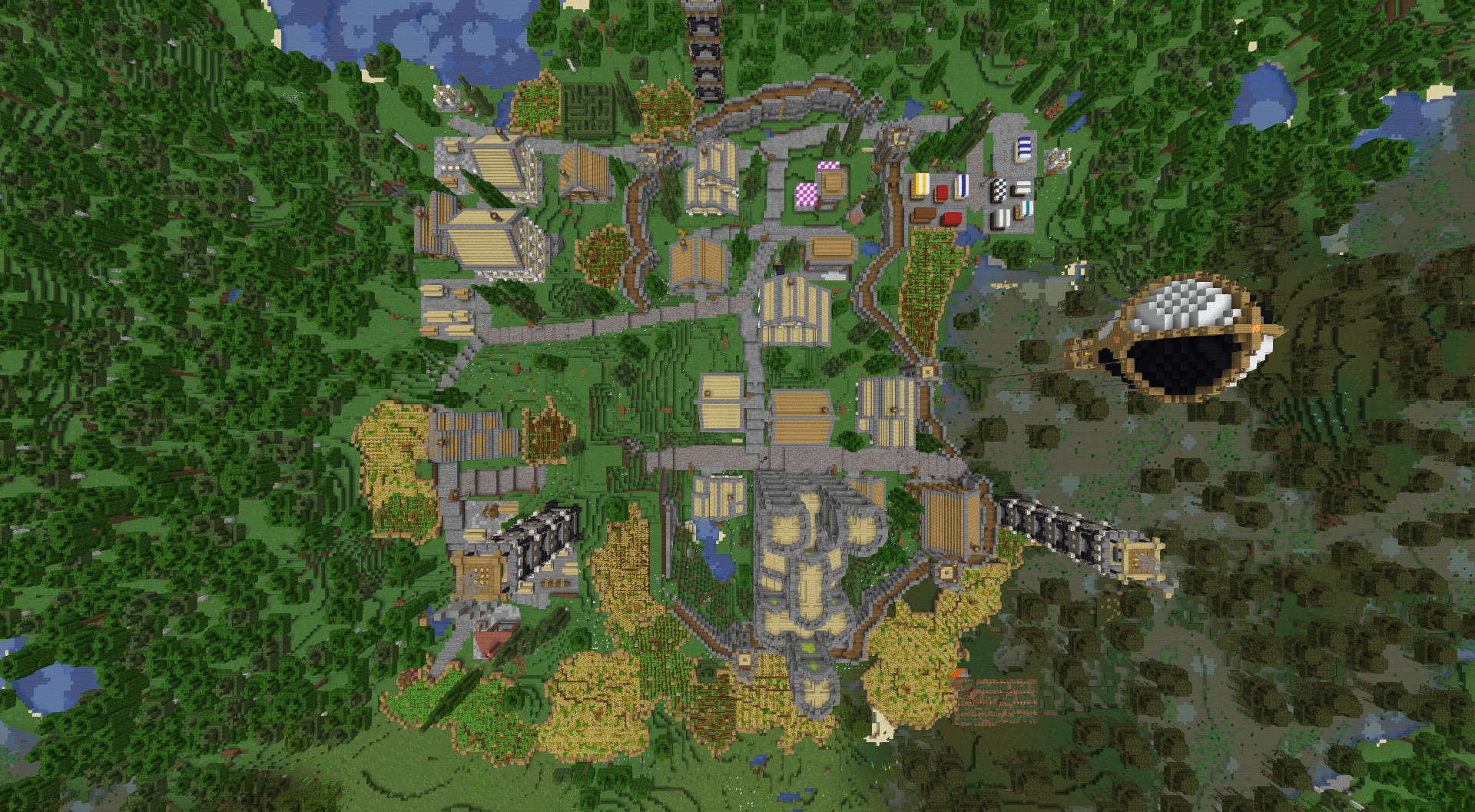}
		\caption{Top-down view of the city in Figure~\ref{fig:birdseye_plains}}
	\end{subfigure}
	\hfill
	\begin{subfigure}[b]{0.48\linewidth}
		\includegraphics[width=\linewidth]{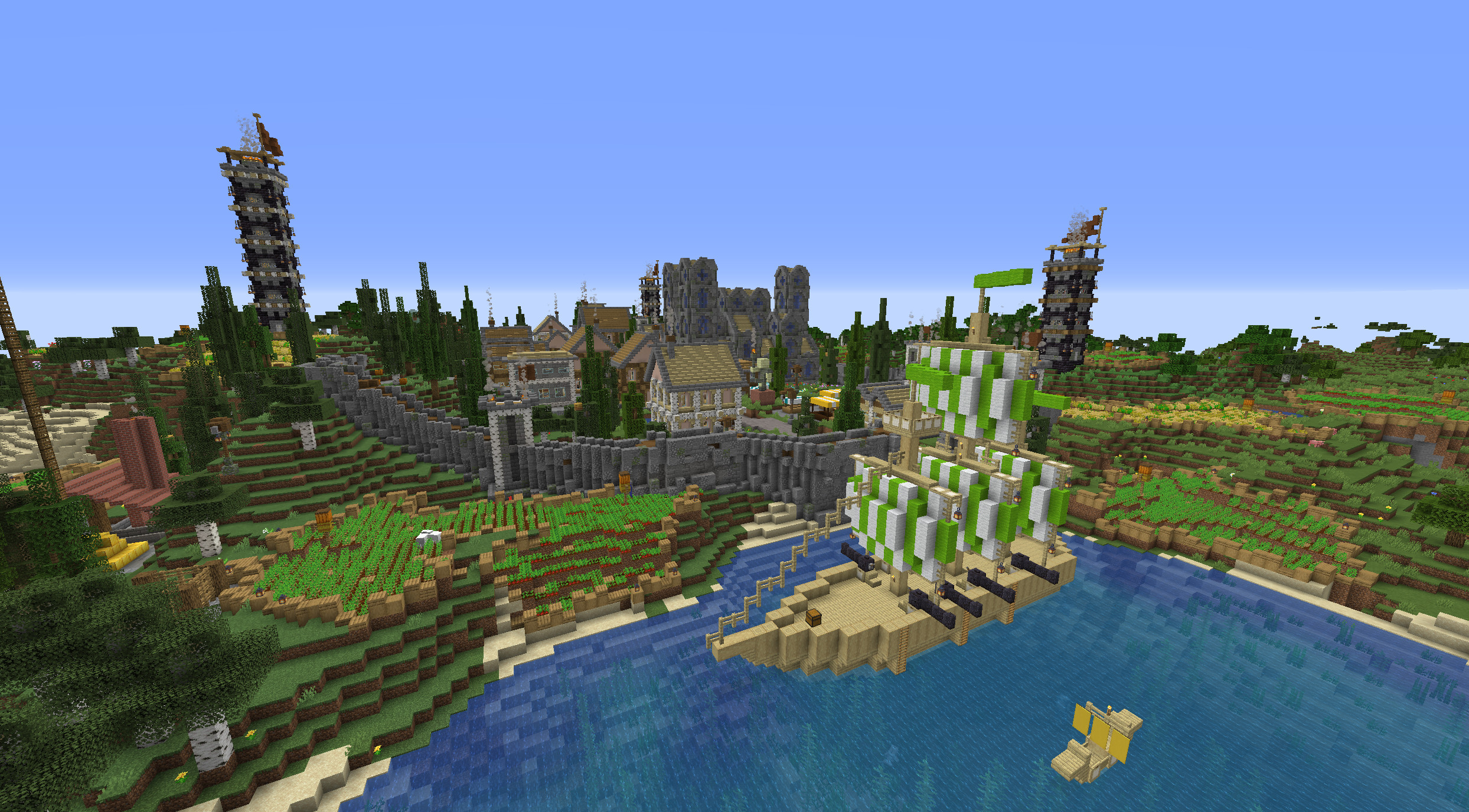}
		\caption{View of a waterside city.}
	\end{subfigure}
 	\hfill
    \begin{subfigure}[b]{0.48\linewidth}
		\includegraphics[width=\linewidth]{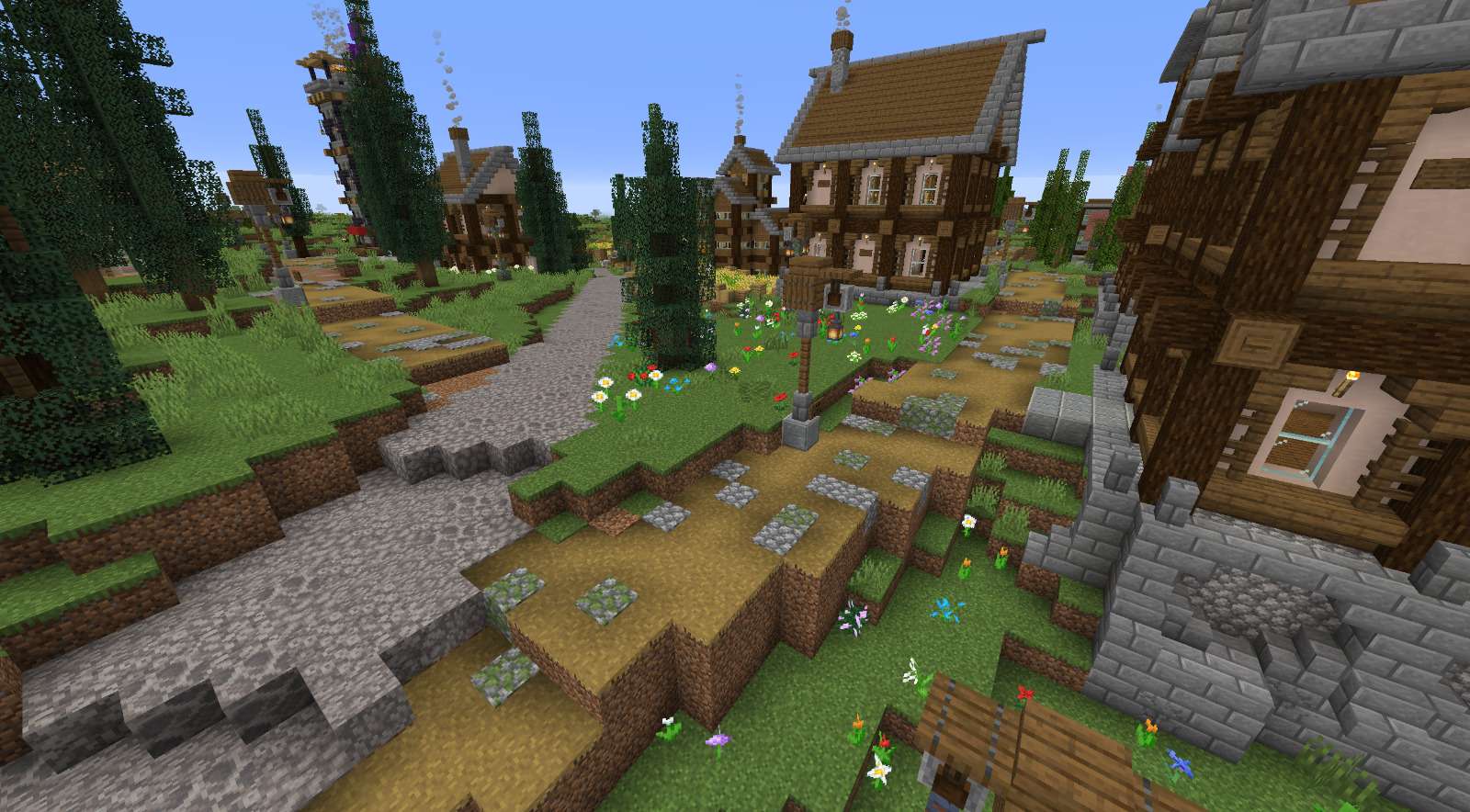}
		\caption{View of a splitting of three streets.}
	\end{subfigure}
    \caption{General impressions of the generated settlements.}
    \label{fig:city_impression}
\end{figure*}

\subsection{Building Differentiation Phase}
During the planning process, plots of different types are determined. Some of these have a defined single purpose, like a watchtower or a boat. However, we also generate more generic \textit{civilian plots}, which just indicate a good road-connected location for a building, but do not specify exactly what type. In the building differentiation phase, we separate these into three distinct building types, as also described by Lechner~\cite{Lechner2003ProceduralCM}, namely: residential, commercial and industrial.

This differentiation is done by another agent-based simulation. Three agents, one for each building type, try to convert a generic plot into a specific plot of their own type in each simulation step. To convert a plot, they must first meet their constraints, and then choose precisely which plot they will convert. 

We dynamically model the constraints as a fictive economy of workers, goods and income. The residential agent adds workers to the population as long as unemployment is below a certain threshold. The industrial agent requires workers, but creates goods and income. The commercial agent requires workers, goods and money. This naive economic interaction leads to a certain balance of all three building types.

If an agent meets all its constraints, it can choose a plot. This choice is based on a fitness function that will take the other surrounding building types into account. For example, a commercial building \q{likes} being close to residential and other commercial buildings but wants to be far away from industry. This leads to some emergent properties like clustering.

\subsection{Block Placement Phase}
Once the final blueprint is finished, the block placement within Minecraft can begin. This is done one structure at a time. For each structure, a church for example, we define a procedural generator that builds the structure within the confinements of the planned-out plot. The level of PCG can vary greatly between structure types, from simple palette swaps up to completely random buildings with different shapes and interiors. Software development can also be parallelised since the procedural generator of a structure is separate from the agent that decides the structure's place in the blueprint.
We describe some of the individual structures in more detail in sect.~\ref{sec:structures}.
Figure~\ref{fig:city_impression} gives a general impression of the generated cities once built in Minecraft.

\subsection{Written-form Narrative}
One criterion of the GDMC competition is the narrative aspect that a city presents. In our approach, written-form narrative is a strong driver of this. Each city gets its own name, street names and names of persons. We used a Markov chain that was modelled on Dutch names for each of these categories. The model probabilistically switches between 2, 3, 4 and 5-gram sequences when generating new names ($N$-gram here referring to $N$ letters). In this way, we switch between more novelty (2,3-grams) and more stability (4,5-grams) and get the benefit of both.

Minecraft supports the creation of in-game books. This can be used to add written-form narrative to a city. The GDMC competition has capitalised on this by introducing a \q{chronicle challenge}~\cite{salge2019chronicle}.  For our submission, we opted for a city chronicle told from the perspective of the mayor. The chronicle goes through civilian buildings in chronological order. This is possible due to the iterative nature of the blueprint planning phase. It mentions their address and build year, and then presents its inhabitants. We randomly generate inhabitants with a name, age, sex, address, marital status, and possibly children. The prose is written with string substitutions and randomly chosen synonymous chunks for the sake of variety (e.g. \q{her business is not doing so well due to [a rat infestation / lacking supplies]}. We also used the same approach for writing a captain's log to describe the contents and (fictive) journey of a merchant ship, if the city generates one.

\section{Blueprint Planning Agents}
\label{sec:agents}

The specifics of the city blueprint are dependent on what agents are used in the iterative expansion process, how many, and what parameters they use. It is, for example, possible to have two agents that place trees but that differ purely in terms of parameters: one places trees inside the city walls once every five steps, the other outside the walls every two steps. Possible parameters are given in Table~\ref{table:plot_agent_parameters}. Next, follows an overview of the types of agents that comprise our own implementation:

\smallskip

\begin{compactdesc}
    \item[3 road agents] to split existing roads in two, to add new roads to decrease $A^{*}$ travel distance between random points in the city, and to bridge over water.
    \item[1 road agent per road] tasked with expanding only that particular road.
    \item[1 wall agent] activates only once and places the city wall.
    \item[3 civilian plot agents] place ``generic civilian plots'' later to be differentiated. We use 3 of them, simply to increase their activation frequency.
    \item [5 farm agents] place different farm types (e.g. wheat, potatoes).
    \item [9 tree agents] activate after the wall step. Their quantity simply increases their activation frequency and creates a green city.
    \item [3 water plot agents] place boats (small and large) and fishing platforms.
    \item [11 agents] for small decorative structures (e.g. wheelbarrows, benches).
    \item [1 agent] for watchtower placement.
    \item [1 agent] for church placement.
\end{compactdesc}

\begin{figure}
    \centering
    \small
    \begin{tabularx}{\columnwidth}{>{\raggedright\arraybackslash}p{2.3cm}X}
        \toprule
        \textbf{Parameter} & \textbf{Explanation \& Example use} \\
        \midrule
        Activation step & Step at which the agent turns active. The default is 0, but structures that depend on the \emph{walled-in map} will want a value larger than the \emph{wall step}.\\
        \addlinespace
        Activation \newline interval & How often the agent is active and tries to perform an action. The default is 1 (i.e. every step). A setting of 2 will activate the agent every other step. \\
        \addlinespace
        Inside-wall-only, \newline Outside-wall-only & Whether the plot is restricted to building only inside/outside of the city walls. \\
        \addlinespace
         Maximum plots & Maximum number of plots of the same kind. For example, limits a city to having only one church. \\
        \addlinespace
        Max distance from roads & How far away from a road this plot can still be built. \\
        \addlinespace
        Water-only & For water structures like boats. \\
        \addlinespace
        Critically important & Example: church. enables bulldozing of other buildings. \\
        \addlinespace
        Distance to own kind & Used when evaluating potential plot locations. A minimum or maximum distance between plots of the same kind. Either to prevent or create clustering. It is also possible to set a weight multiplier (positive or negative), instead of a hard cut-off distance. \\
        \addlinespace
        Slope & Used when evaluating potential plot locations. Both a maximum allowed slope as well as a weight multiplier. \\
        \addlinespace
        Custom evaluation function & Gives a score to a potential plot location based on a designer-specified requirement not listed above. What is good or bad highly differs between agent specifications.  \\  
        \bottomrule
    \end{tabularx}
    \captionof{table}{Example parameters that can be used by agents in the blueprint planning phase.}\label{table:plot_agent_parameters}
\end{figure}

\section{Structures}
\label{sec:structures}

Although we focus mainly on the overall agent-based blueprint simulation, some of the individual structures that are created in the block placement phase use some notable PCG techniques as well. Nearly all structures employ some form of randomised or terrain-adaptive variation. Here, we describe some of the more complex structures.

\subsection{Cell structures}
\begin{figure}[H]
    \centering
	\includegraphics[width=\columnwidth]{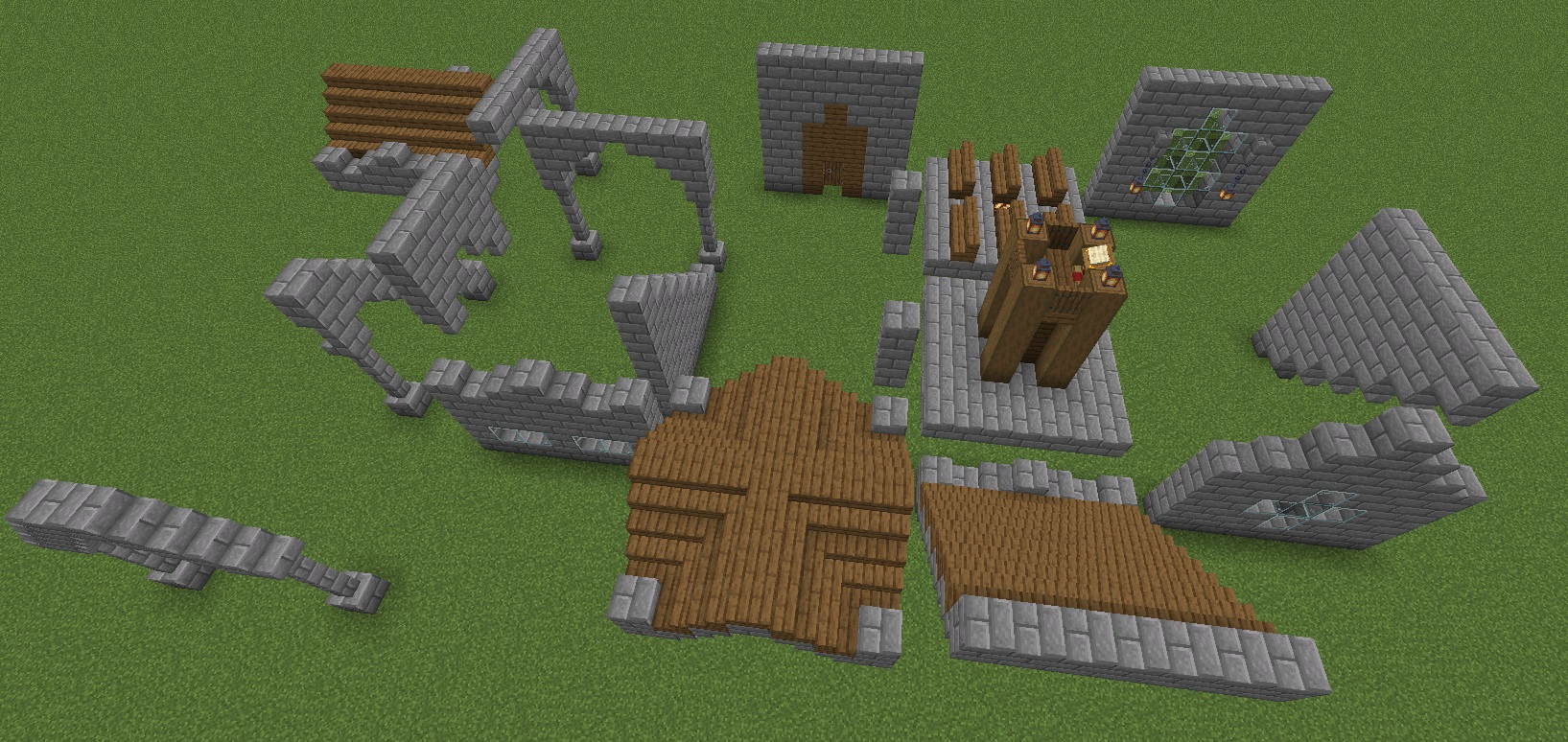}
	\caption{Overview of the prefabricated models that combine to form the church.}
    \label{fig:church_models}
\end{figure}

Some structures, such as the church and some of the residential houses, are generated by combining prefabricated models (see Figure~\ref{fig:church_models}). All models are the same size so that they can be tiled as square cells. A heightmap then encodes the shape of the structure. In the case of the church, for example, a cell of height 8 surrounded by cells of a much lower height represents a tower. Thus, by looking at the neighbours of each cell, we can determine which of the models to use. By randomly modifying a heightmap, we add variation to the final result. This method is also not limited to rectangular shapes.

\subsection{City wall}
To promote a sense of history and to give the city some additional structure, we incorporated a city wall. The wall is tightly integrated with the agent-based simulation: while most structures are only roughly positioned in the blueprint planning phase and then made concrete in the block placement phase, the wall is already planned out in great detail during the blueprint phase. Like the roads, it can follow any arbitrary curve and is not limited to rectangles.

The wall agent activates only once during the blueprint planning phase, at which point it defines a wall that surrounds all existing building plots at that point. This way, we leverage the organic nature of our simulation system to give the wall a realistic shape. The exact wall points are computed by determining the convex hull of all existing plots, pushing these convex hull points outwards, connecting the points using weighted A* pathfinding, and widening the resulting band of points.

In the block placement phase, the band of points is used to realise the wall. Its rugged, organic design makes it adaptable to any path. The top surface of the wall is kept stable and walkable by smoothening the heights at each point using a Savitzky-Golay filter, and the wall's uniformity is broken with some periodic wall towers.

\subsection{Farms}

A large amount of ground outside the city wall is filled with organically-shaped farms that follow the contours of the original terrain. We use the same algorithm to create farms with various kinds of crops, such as wheat and potatoes. Farms are unique in that they can expand outside their assigned plot in the block placement phase. This allows them to closely hug other structures and roads in ways that would not be achievable with rectangular plots.

The expansion algorithm for farms consists of a number of steps. The first step is the placement of a few initial farm \q{blobs} in the originally assigned plot. Then, for each block height level, the farm points at that level are morphologically dilated using a mask of points that are at the same height and that are not reserved for another structure. This per-height expansion is what causes farms to follow terrain contours.

After all individual farms have been expanded and built, we build borders around them. Placing borders after expanding all farms allows adjacent farms to merge into larger ones, providing some additional size variation.

\section{Performance}
\label{sec:runtime}

We may wonder how long the whole generation process actually takes, and which components of it are computationally most expensive. Therefore, we perform an experimental runtime analysis for different map sizes and biomes (i.e. biological regions in Minecraft, like deserts or forests). 

We investigated three parameters: the number of time steps in the blueprint planning phase, the size of the build area, and the biome(s) of the build area. To get a general sense of the runtime, we conducted experiments in a Minecraft \emph{single-biome world} (a custom world setting). We set the biome to plains and experimented with the number of blueprint time steps and build area size. All experiments were run on an AMD Ryzen 7 3700U processor (2.3 GHz).

\begin{figure}
    \centering
	\includegraphics[width=\columnwidth]{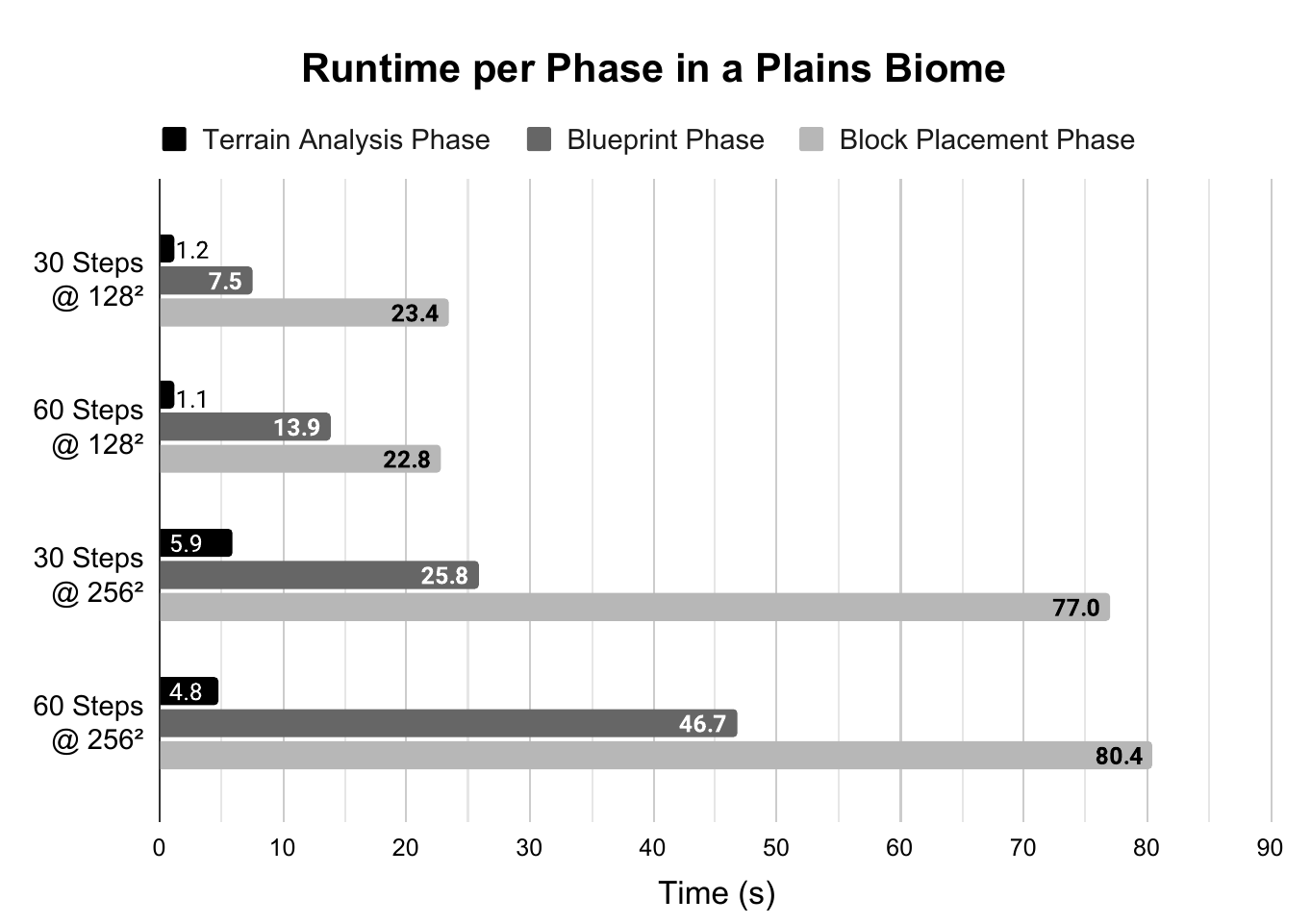}
	\caption{Runtime of the algorithm in a plains biome split into 3 phases, while varying the number of blueprint steps (30 or 60) and build area size (128x128 or 256x256). Average of 5 runs per parameter.}
    \label{fig:runtime_plains}
\end{figure}

The total runtimes are shown in Table~\ref{table:total_runtime}. In Figure~\ref{fig:runtime_plains}, these runtimes are shown split up by their phases. Note that here, we have included the building differentiation phase in the \emph{blueprint phase} in the figures, as both make edits to the blueprint. We can see that the size of the build area is the most significant factor. Additionally, the block placement phase does not necessarily take longer if we double the number of blueprint steps, presumably because the last 30 steps do not add as much as the first 30 steps (as seen in Figure~\ref{fig:blueprint_evolution}).

\begin{figure}
    \centering
    \textbf{Total Runtime (s)} \\
    \smallskip
    \begin{tabular}{lrr}
        \toprule
        \multicolumn{1}{l}{\textbf{Build Area}} &
        \multicolumn{2}{c}{\textbf{Blueprint steps}}    \\ 
        \cmidrule(lr){2-3} &
        \multicolumn{1}{r}{\textbf{30}} &
        \multicolumn{1}{r}{\textbf{60}}     \\
        \midrule
        \textbf{128×128} & 33.0 & 38.7 \\
        \addlinespace
        \textbf{256×256} & 111.7 & 134.3 \\
        \bottomrule
    \end{tabular}
    \captionof{table}{The total runtime in a plains biome. Average of 5 runs per parameter.}\label{table:total_runtime}
\end{figure}

\begin{figure}
    \centering
	\includegraphics[width=\columnwidth]{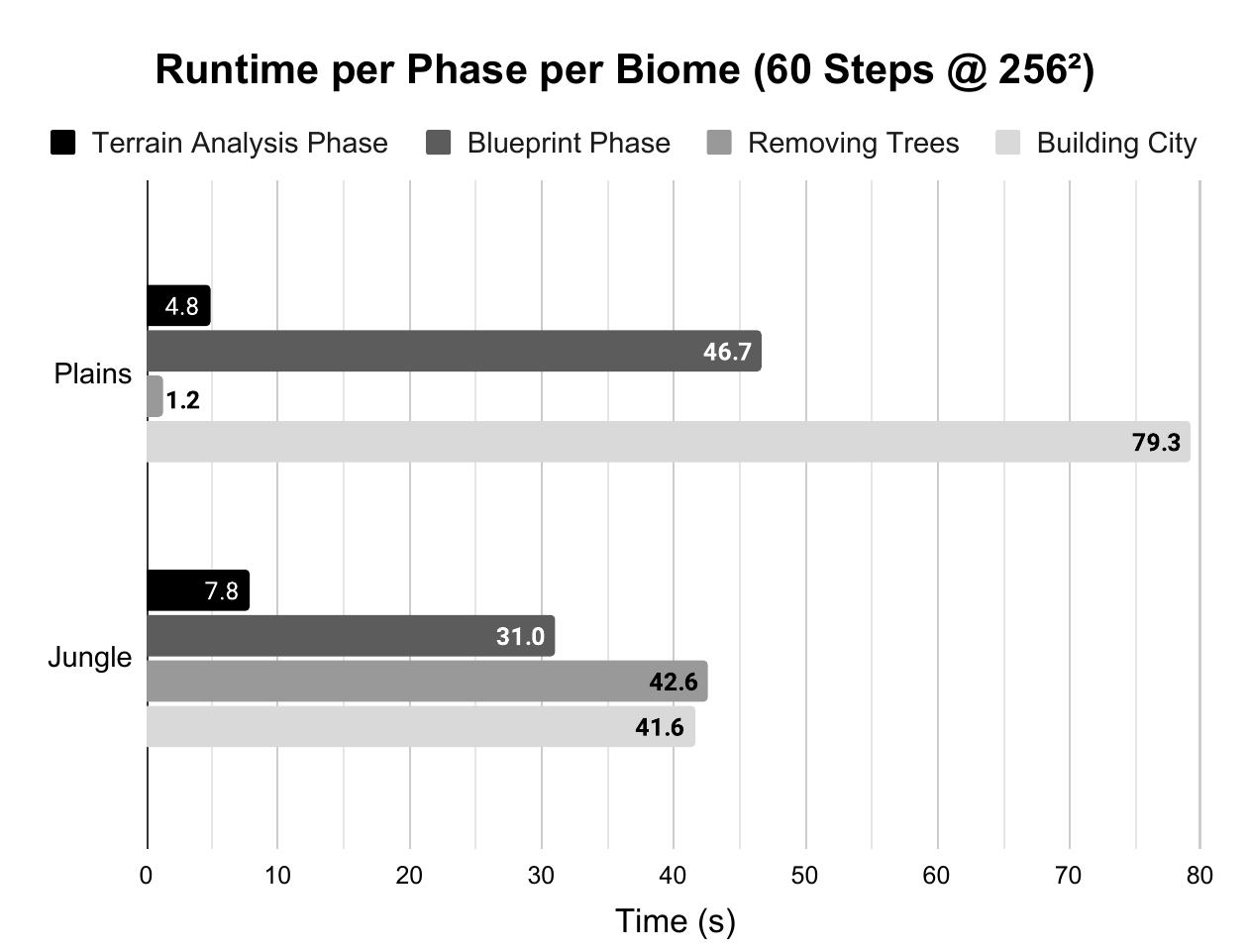}
	\caption{Runtime of the algorithm in a plains and jungle biome, with the block placement phase split into preparatory tree removal and actual city building. 60 blueprint time steps were used in a 256x256 build area.}
    \label{fig:runtime_jungle_plains}
\end{figure}

Next, we compared the plains biome to the jungle biome. These biomes represent two extremes: plains with little or no trees and a jungle filled with tall trees and bushes. We separated the runtime of the tree removal in the block placement phase from the rest of the phase. The results in Figure~\ref{fig:runtime_jungle_plains} suggest that removing trees in the jungle is an expensive operation. The jungle terrain presents a challenge to the blueprint planning phase: agents cannot always perform their actions due to a lack of space. 

With less city development, the total runtime decreases. The average total runtime in a plains biome (256×256, 60 blueprint time steps) was 134.3 seconds. In the jungle, this was 125.8 seconds. The latter, however, is an average of four runs. A fifth was found to be a large outlier (350.3 seconds, 8$\sigma$ from the mean), and produced a much larger city than the other four, presumably due to favourable terrain by chance.

\section{Jury evaluation}
\label{sec:evaluation}

All GDMC submissions are judged by a human jury in four categories: \textit{Adaptability}, \textit{Functionality}, \textit{Evocative narrative} and \textit{Aesthetics}. 
The jury members assign a grade from 0 to 10 (10 being the best) for each category. More details about these categories can be found in~\cite{salge2018generative}. Notably, the grades are given on an absolute scale, making it possible to compare the scores of all submissions across all the competitions that have been held so far. For each category, a score of 5 is meant to be given if a submission performs at the level of a \q{naive human}, and a score of 10 would indicate \q{superhuman performance}~\cite{gdmc-wiki-evaluation}.
An analysis of the more granular themes that the GDMC judges tend to base their scores on has also been performed by Hervé et al.~\cite{Herve2023}.

There currently exist no satisfactory formal evaluation criteria for procedurally generated Minecraft settlements. In fact, such criteria are lacking in most areas of computational creativity~\cite{salge2018generative}. Although human evaluations can be imprecise, this lack of formal criteria nonetheless makes the human evaluations provided by the GDMC competition judges a valuable resource for the evaluation of new approaches.

\begin{figure*}
    \centering
    \subcaptionbox{Adaptability}{\includegraphics[width=0.24\textwidth]{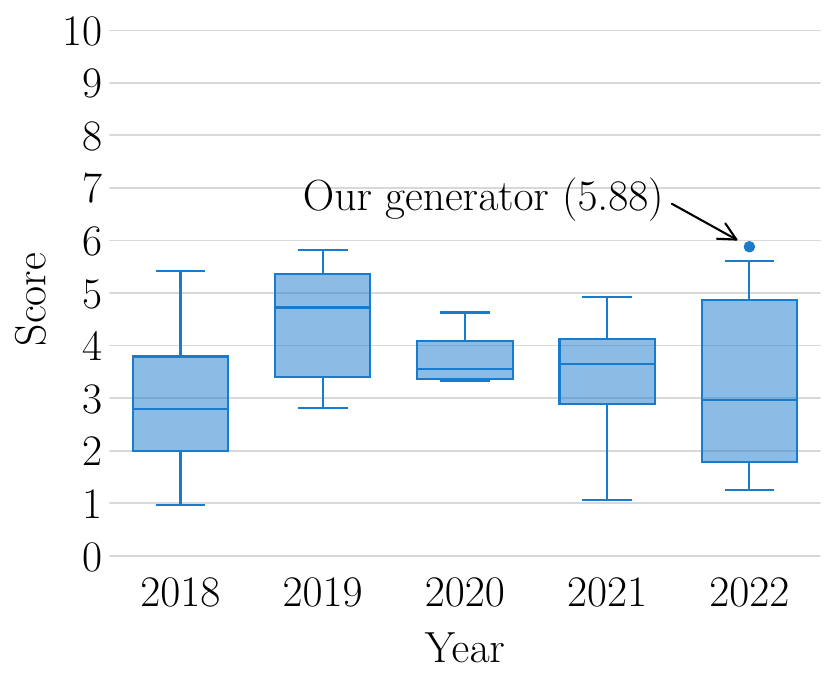}}
	\subcaptionbox{Functionality}{\includegraphics[width=0.24\textwidth]{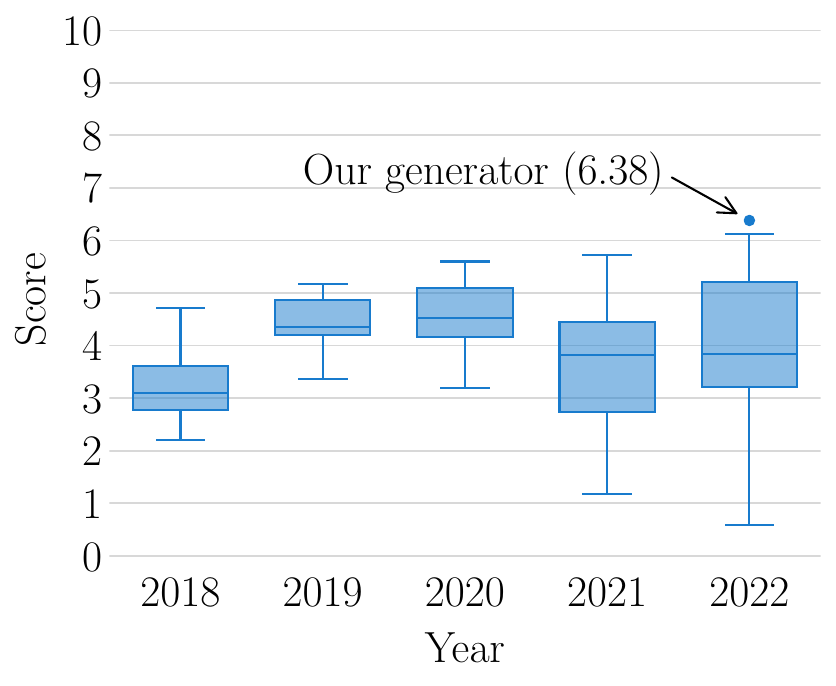}}
 	\subcaptionbox{Narrative}{\includegraphics[width=0.24\textwidth]{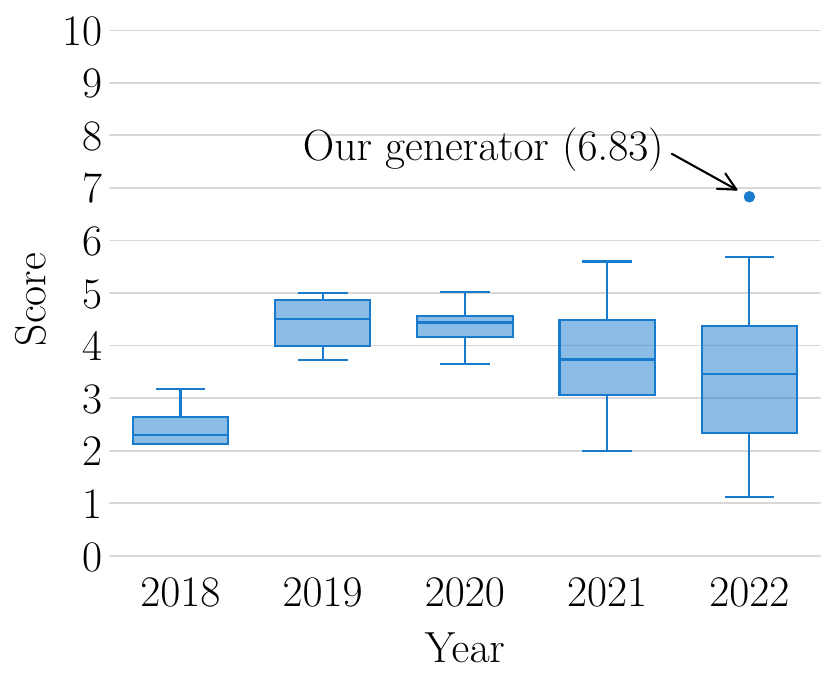}}
  	\subcaptionbox{Aesthetics}{\includegraphics[width=0.24\textwidth]{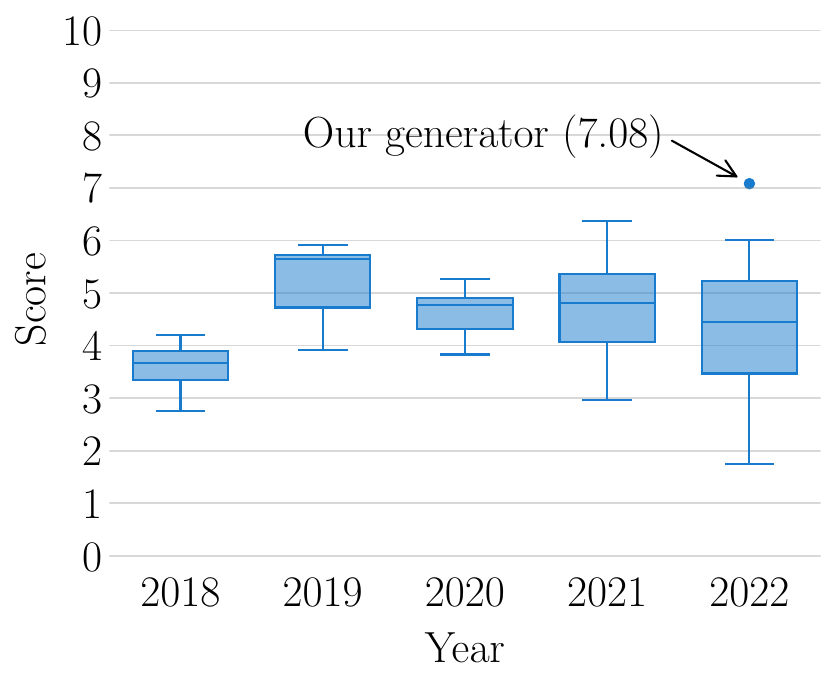}}
	\caption{Overview of the scores that have been achieved in each scoring category of the GDMC competition throughout the years, as awarded by the judges. The scores are grouped by competition year. Note that our own scores are not included in the 2022 boxplots. Our generator achieved the highest historic score in every category.}
    \label{fig:gdmc-scores}
\end{figure*}

Figure~\ref{fig:gdmc-scores} shows a comparison between the performance of our city generator and all other GDMC submissions, in regard to the four GDMC scoring categories. Our generator has managed to achieve the highest historic score in every category, showing the effectiveness of our iterative approach. It achieved particularly high scores in the categories \textit{Narrative} (6.83) and \textit{Aesthetics} (7.08), greatly improving on the previous bests. Achieving a high score in all categories at once also indicates that our approach is effective at properly balancing the various facets of Minecraft generative settlement design.

\section{User Study (Jury)}
\label{sec:user-study}

Instead of doing our own user study, we look at the 
comments received from the Jury of the GDMC 2022 competition. Even with such a small sample, this enables us to see how the method described herein performed in comparison to other recent approaches. 

Our submission to the GDMC competition received feedback from eight competition judges. The city was found to be aesthetically pleasing, with a great variety of buildings. One judge specifically noted the modular variation in some buildings. The generator also gave the impression of a lived-in settlement; the city felt logical with a strong sense of narrative. Lastly, two judges noted that the city walls worked well with following the contours of the terrain. Criticism was given as well. The written chronicle is currently too verbose, and building placement may feel insufficiently adapted to the terrain (e.g. floating foundation platforms). One judge noted that the furnishing of buildings can seem odd. Lastly, building materials could have been better adapted to the environment.

\section{Future Work}
\label{sec:future-work}

Some possible extensions to the generation process could include making the list of active agents more adaptive. For example, in a mountainous region, we could add tunnel-digger or mine-digger agents. We could also set some more conditions, such as some buildings only being able to be built once the city reaches a certain size, or the opposite, with certain plots being removed if the city becomes too large. 

For creating the structures in the block placement phase, we could also add some improvements. We could mimic a historical city centre if the block placement phase uses styles of different time periods depending on when the blueprint planning phase developed a plot. Cities could also be given a cultural theme depending on the biome. Desert cities could, for example, be given an Egyptian style and taiga cities a Viking style. This could affect both the block placement phase (in the form of a different architecture, or building materials as suggested by the jury) and the blueprint planning phase (e.g. only building farms next to rivers to mimic Nile Valley civilisations).

Another aspect that could be improved, as noted by the jury, is the furnishing of the structures. Although we furnish houses with a good variety of furniture items, the choice and placement of these items is not particularly sophisticated. For example, houses may generate without a bed or may contain many duplicates of the same item in one room.

Furthermore, buildings currently do not really \q{acknowledge} each other. To fix this, we could introduce a new phase between the blueprint and block placement phases which seeks to add some connections between structures: making them share an outer wall, making walking bridges between buildings or simply hanging clothing lines between them. We believe this would increase the cohesion of structures and could lead to some more interesting and varied generations and might make the overall city more believable.

Another avenue of improvement would be to plan the blueprint out in 3D, rather than just in 2D. Our current approach is unable to plan out structures that overhang others. Adding this capability could also help to improve cohesion, especially for supporting structures like trees.

Finally, we would also like to perform a quantitative analysis of generated settlements in order to quantify the range of the generative space. This could be done with metrics like city size, building density, building sizes, or building variation. But since settlement generation is a subjective challenge, jury scores are a more valuable measure.

\section{Conclusion}
\label{sec:conclusion}

In the GDMC competition 2022, every single jury member rated the herein described submission as their rank 1 entry, making it the winner. But what actually makes it that good? We have provided an overview of the general concept and how it enables a parallel software development process with five developers. We have also provided some feedback obtained from our first \q{users}, the competition jury. In our view, the flexibility of the approach that \q{unleashes} the creativity of the developers by means of the agent-based modular planning and building systems is the main reason for the success. This approach can easily be transferred to slightly or completely different settings by adding or replacing some or all of the blueprint planning agents as envisioned in sect.~\ref{sec:future-work}.

\bibliographystyle{IEEEtran}
\bibliography{IEEEabrv, bibliography/bibliography.bib}
\end{document}